\documentclass[a4paper]{article}

\usepackage{INTERSPEECH2016}

\usepackage{graphicx}
\usepackage{amssymb,amsmath,bm}
\usepackage{textcomp}

\usepackage{xcolor}

\ninept

\title{Sequential Convolutional Neural Networks for Slot Filling in Spoken Language Understanding}

\makeatletter
\def\name#1{\gdef\@name{#1\\}}
\makeatother \name{{\em Ngoc Thang Vu}}

\address{Institute of Natural Language Processing, University of Stuttgart \\
  {\small \tt thangvu@ims.uni-stuttgart.de}
}

\begin{document}

  \maketitle
\begin{abstract}
We investigate the usage of convolutional neural networks (CNNs) for the slot filling task in spoken language understanding.
We propose a novel CNN architecture for sequence labeling which takes into account the previous context words with preserved order information and pays special attention to the current word with its surrounding context.
Moreover, it combines the information from the past and the future words for classification.
Our proposed CNN architecture outperforms even the previously best ensembling recurrent neural network model and achieves state-of-the-art results with an F1-score of 95.61\% on the ATIS benchmark dataset without using any additional linguistic knowledge and resources.
  \end{abstract}

 \section{Introduction}
The slot filling task in spoken language understanding (SLU) is to assign a semantic concept to each word in a sentence. 
In the sentence \textit{I want to fly from Munich to Rome}, an SLU system should tag \textit{Munich} as the departure city of a trip and \textit{Rome} as the arrival city. 
All the other words, which do not correspond to real slots, are then tagged with an artificial class \texttt{O}.
Traditional approaches for this task used generative models, such as hidden markov models (HMM) \cite{hmm_1}, or discriminative models, such as conditional random fields (CRF) \cite{crf_1, crf_2}. 
More recently, neural network (NN) models, such as recurrent neural networks (RNNs) and convolutional neural networks (CNNs) have been applied successfully to this task \cite{rnn_slu_1, rnn_slu_2, cnn_slu, lstm_slu, rnn-mem}. 

Overall, RNNs outperformed other NN models and achieved the state-of-the-art results on the ATIS benchmark dataset \cite{rnn-rank}.
Furthermore, bi-directional RNNs have worked best so far showing that information from both the past and the future is important in predicting the semantic label of the current word.
It is, however, well known that it is difficult to train an RNN due to the vanishing gradient problem \cite{rnn-prob}.
Introducing long short-term memory (LSTM) \cite{lstm} or other variants of LSTM such as the gated recurrent unit (GRU) can solve this problem but, in turn increases the number of parameters significantly.
Previous results reported in \cite{rnn-mem} did not show any improvement on the ATIS data set using LSTM or GRU.

In contrast to previous papers which reported state-of-the-art results with RNNs, we explore the usage of convolutional neural networks for a sequence labeling task like slot filling. 
Previous research in \cite{cnn_slu} showed promising results on the slot filling task.
The motivation behind this is to allow the model to search for patterns in order to predict the label of the current word independent of the feature representation of the previous word.
%Furthermore, it does not force the model to store the previous content information into a fix length vector.
Moreover, CNNs provide several advantages: it preserves the word order information, it is faster and easier to train and does not mix up the word sequence and therefore it is able to interpret the features learnt for the current task to some extent.

This study investigates the usage of CNNs for a sequential labeling task like slot filling with the following contributions:

(1) We propose a novel CNN architecture for sequence labeling which takes into account the previous context words with preserved order information and pays special attention to the current word with its surrounding context.

(2) We extend the proposed CNN model to a bi-directional sequential CNN (bi-sCNN) which combines the information from past and future words for prediction.

(3) We compare the impact of two different ranking objective functions on the recognition performance and analyze the most important n-grams for semantic slot filling.

(4) On the ATIS benchmark dataset, the proposed bi-directional sequential CNN outperforms all RNN related models and defines a new start-of-the-art F1-score of 95.61\%.

%The remainder of the paper is organized as follows:
%Section~\ref{sec:related} gives an overview of related works.
%We describe the proposed CNN architecture and the learning objective functions in Section~\ref{sec:cnn}.
%Section~\ref{sec:results} presents the experimental results on the ATIS benchmark data set.
%In Section~\ref{sec:comparison}, we compare the proposed CNN with previous state-of-the-art results and summarize the work in Section~\ref{sec:conclusions}.

\section{Related Work}
\label{sec:related}
Neural network models such as RNNs and CNNs have been used in a wide range of natural language processing tasks. 
Vanilla RNNs or their extensions such as LSTMs or GRUs showed their success in many different tasks such as language modeling \cite{rnn_lm} or machine translation \cite{translation}.  
Another trend is to use convolutional neural networks for sequence labeling \cite{nlp_1,nlp_2} or modeling larger units such as phrases \cite{phrase} or sentences \cite{sent_1, sent_2}.
For both models, distributed representations of words \cite{word_1, word_2} are used as input.

In the spoken language understanding research area, neural networks have also been applied to intent determination or semantic utterance classification tasks \cite{dnn_slu_1, dnn_slu_2}.
For the slot filling task, RNNs \cite{rnn_slu_1, rnn_slu_2} and their extensions \cite{lstm_slu, rnn-mem} outperformed not only traditional approaches but also other neural network related models \cite{cnn_slu} and defined the state-of-the-art results on the ATIS benchmark data set.
Recently it was shown in \cite{rnn-rank} that applying ranking loss to train the model is effective for tasks that involve an artificial class like \texttt{O}.
They achieved state-of-the-art F1-scores of 95.47\% with a single model and 95.56\% by combining several models. 
In summary, the RNNs appear to be the best model for this task to date.
The only previous study using convolutional neural networks was presented in \cite{cnn_slu} showing promising results.
However, it did not outperform the RNN related models.

\section{Bi-directional Sequential CNN}
\label{sec:cnn}
\begin{figure}
\includegraphics[width=.45\textwidth]{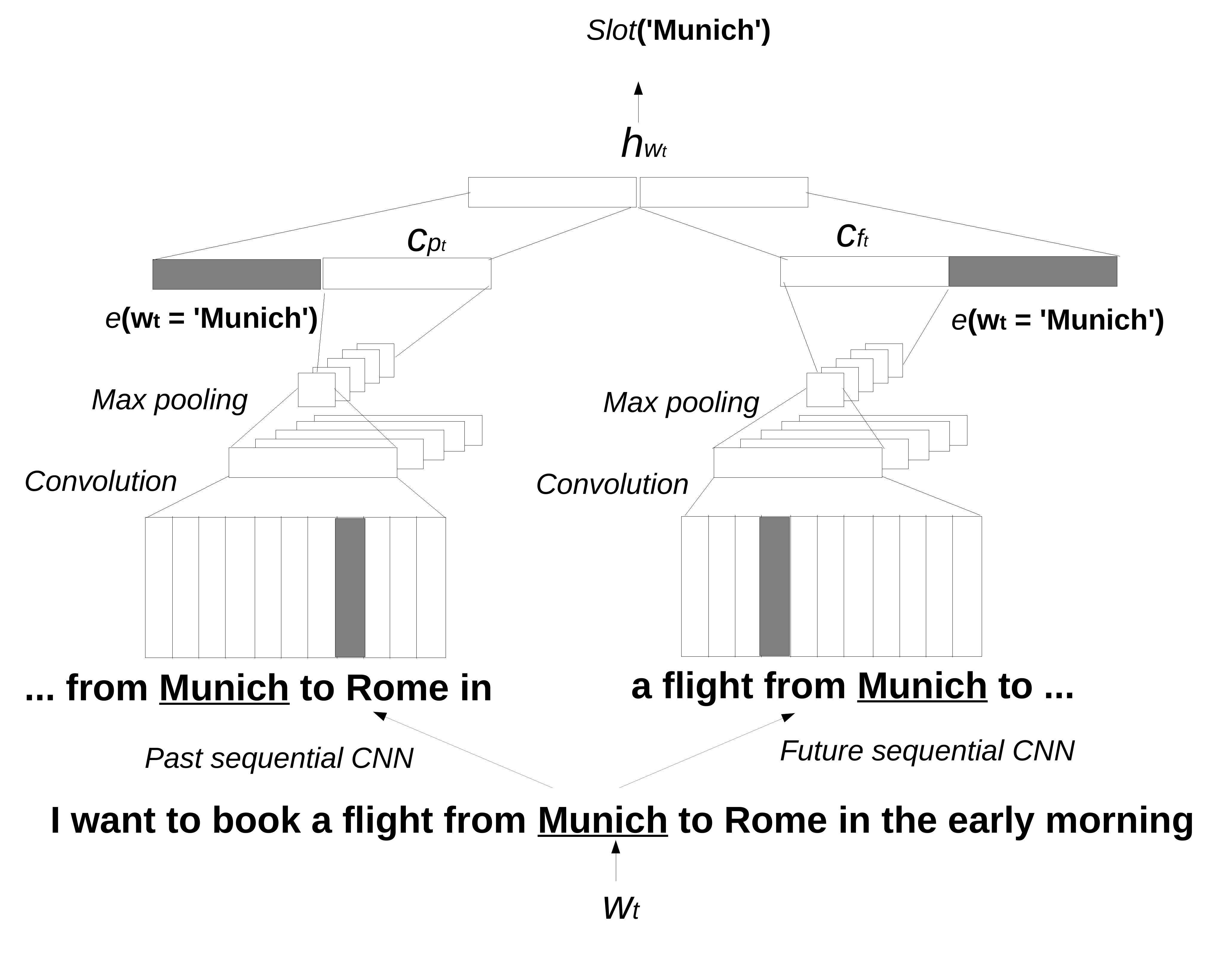}
\caption{Bi-directional sequential CNN (bi-sCNN) which combines past and future sequential CNNs for slot filling}
\label{fig:biCNN}
\end{figure}

This section describes the architecture of the bi-directional sequential CNN (bi-sCNN) illustrated in Figure \ref{fig:biCNN}.
It contains three main components: a vanilla sequential CNN, an extended surrounding context and a bi-directional extension.
\subsection{Model}
\textbf{Vanilla sequential CNN.}
To predict the semantic slot of the current word $w_t$, we consider $n$ previous words in combination with the current word.
In order to avoid the border effect, the $m$ future padding words are also included.
Each of the words is embedded into an $d$-dimensional word embedding space.
Thus for each current word, we form a matrix $w \in R^{(n+m+1) \times d}$ as an input to the CNN for prediction.

There are several possibilities for convolving the input matrix: applying 1D filters to each dimension independently or applying 2D filters spanning some or all dimensions of the word embeddings. 
In this paper, we use 2D filters $f$ (with width $|f|$) spanning all embedding dimensions $d$.
This is described by the following equation:
\begin{equation}
 (w \ast f)(x,y) = \sum_{i=1}^{d}\sum_{j = -|f|/2}^{|f|/2}w(i,j) \cdot f(x-i,y-j)
\end{equation}
where $w$ is the word matrix and $f$ is the filter matrix.
On each output, a nonlinear function such as the sigmoid function can be applied.
After convolution, we use a max pooling operation to find the most important features.
This function stores only the highest activation of each convolutional filter for the succeeding steps.
If $s$ filter matrices are used, an $s$-dimensional feature representation vector $c_{p_t}$ is created for further classification.

\textbf{Extended surrounding context.}
When moving from one word to the next, the input matrix changes only slightly which leads to a large overlap of detected features from the convolutional and max pooling operator.
Furthermore, the model needs to know which word is the current word for slot prediction.
Therefore, in order to pay special attention to the current word and use the information of the word itself directly for the prediction, we introduce an additional component which uses the current word and its surrounding context words as input vector $e(w_t)$ with $d(2*cs+1)$ dimensions.
$cs$ is the surrounding context length.
The feature representation of the current word is computed as follows:
\begin{equation}\label{surrounding}
\begin{split}
h_{w_t} = f(U \cdot e(w_t ) + V_p \cdot c_{p_t})
\end{split}
\end{equation}
where  $U \in R^{s \times d(2*cs+1)}$ and $V_p \in R^{s \times s}$.

\textbf{Bi-directional sequential CNN.}
As reported in \cite{rnn-rank}, information not only from the past but also from the future contributes to the recognition accuracy.
We therefore extend the sequential CNN to the future context. 
Because CNN preserves order information, we do not scan the input text from right to left like a bi-directional recurrent neural network. 
Instead, we take $n$ future words in combination with the current word and the $m$ previous padding words in the original order to form a matrix  $w \in R^{(n+m+1) \times d}$ as an input to the future sequential CNN.
Convolutional and max pooling operators are applied as in the vanilla sequential CNN to obtain a feature representation vector $c_{f_t}$ for the future context information.

There are two different ways to combine the information from the past and future contexts.
The combination can be achieved by a weighted sum of the forward and the backward hidden layer.
This leads to the following hidden layer output at time step $t$:
\begin{equation}\label{bi}
\begin{split}
h_{w_t} = f(U \cdot e(w_t)  + V_p \cdot c_{p_t} + V_f \cdot c_{f_t})\\
\end{split}
\end{equation}
Another combination option is to concatenate the forward and the backward hidden layer.
\begin{equation}\label{bi-add}
\begin{split}
h_{w_t} = [f(U \cdot e(w_t)  + V_p \cdot c_{p_t}),  f(U \cdot e(w_t) + V_f \cdot c_{f_t})]\\
\end{split}
\end{equation}
The combined hidden layer output is then used to predict the semantic label for the current word.
The experimental results in Section \ref{sec:results} show that the combination method is an important design choice that effects the final performance.
\subsection{Training objective function}
\label{ssec:obj}
It was shown in \cite{rnn-rank} that using ranking loss is more accurate than cross entropy to train the model for this task.
One reason might be that it does not force the network to learn a pattern for the \texttt{O} class which in fact may not exist.
In this paper, we compare two different kinds of ranking loss functions.

The first function is the well known hinge loss function:
\begin{equation}\label{hinge-loss}
  L = max(0, 1 - s_{\theta}(w_t)_{y^+} + s_{\theta}(w_t)_{c^-})
\end{equation}
with $s_{\theta}(w_t)_{y^+}$ and $s_{\theta}(w_t)_{c^-}$ as the scores for the target class and the wrongly predicted class of the model given the current word $w$ respectively.
This loss function maximizes the margin between those two classes.

The second one was proposed by Dos Santos et al. \cite{ranking_loss} and used in \cite{rnn-rank} to achieve the current best performance on the slot filling task till now.
Instead of using the softmax activation function, we train a matrix $W^{class}$ whose columns contain vector representations of the different classes.
Therefore, the score for each class $c$ can be computed by using the product
\begin{equation}
 s_{\theta}(w_t)_{c} = h^T_{w_t}[W^{class}]_c
\end{equation}
We use the same ranking loss function as in \cite{rnn-rank} to train the CNNs.
It maximizes the distance between the true label $y^+$ and the best competitive label $c^-$ 
given a data point $x$. 
The objective function is
\begin{equation}\label{obj}
\begin{split}
  L = \log(1+\exp(\gamma(m^+ - s_{\theta}(w_t)_{y^+}))) \\ 
     + \log(1+\exp(\gamma(m^- + s_{\theta}(w_t)_{c^-})))
\end{split}
\end{equation}
with $s_{\theta}(w_t)_{y^+}$ and $s_{\theta}(w_t)_{c^-}$ as the scores for the classes $y^+$ and $c^-$ respectively.
The parameter $\gamma$ controls the penalization of the prediction errors and $m^+$ and $m^-$ are margins
for the correct and incorrect classes.
$\gamma$, $m^+$ and $m^-$ are hyper-parameters which can be tuned on the development set.
For the class \texttt{O}, only the second summand of Equation~\ref{obj} is calculated during training, i.e. the model does not learn a pattern for class \texttt{O} but nevertheless increases its difference to the best competitive label.
Furthermore, it implicitly solves the problem of un-balanced data since the number of class \texttt{O} data points is much larger than in other classes.
During testing, the model will predict class \texttt{O} if the scores for all other classes are $<$ 0. 

\subsection{Comparison with other neural models} 
The information flow of the proposed model is comparable with a bi-directional RNN.
Instead of using the recurrent architecture to save the information from a long context, we use a convolutional operator to scan all the n-grams in the contexts and find the most important features with max pooling.
At every time step, the most important features are then learnt independently from the previous time step.
This poses an advantage over bi-directional RNNs when the previous word is a word of class \texttt{O} and the current word is not of class \texttt{O} because the information to predict class \texttt{O} is not helpful to predict other classes.
Another difference is the integration of future information. 
In the backward RNN model, the sentence is scanned from right to left which is against the nature of languages like English. 
In contrast, the CNN keeps the correct order of the sentence and searches for important n-grams. 

Another interpretation of this model is a joint training of a feed-forward NN and a CNN.
The feedforward NN takes the current word with its surrounding context as input for prediction while the CNN searches for n-gram features from the past and future contexts. 
The context representation of the CNN is used as additional input of the feedforward NN.
This is an advantage of this model over the CNN model proposed in \cite{nlp_2} which has problems identifying the current word for labeling.
 
\section{Experimental Results}
\label{sec:results}
\subsection{Data}
\label{ssec:subhead}
To compare our work with previously studied methods, we report results on the widely used ATIS dataset~\cite{atis_1, atis_2}.
This dataset is from the air travel domain and consists of audio recordings of speakers making travel reservations.
All the words are labeled with a semantic label in a BIO format (B: begin, I: inside, O: outside), e.g. \textit{New York} contains two words \textit{New} and \textit{York} and is therefore labeled with \texttt{B-fromloc.city\_name} and \texttt{I-fromloc.city\_name} respectively. 
Words which do not have semantic labels are tagged with \texttt{O}.
In total, the number of semantic labels is 127, including the label of the class \texttt{O}.
The training data consists of 4,978 sentences and 56,590 words. 
The test set contains 893 sentences and 9,198 words.
To evaluate our models, we used the script provided in the text chunking CoNLL shared task 2000\footnote{http://www.cnts.ua.ac.be/conll2000/chunking/} in line with other related work.
\subsection{Model training}
We used the Theano library~\cite{theano} to implement the model.
To train the model, stochastic gradient descent (SGD) was applied.
We performed 5-fold cross-validation to tune the hyper-parameters. 
The learning rate was kept constant for the first 10 epochs.
Afterwards, we halved the learning rate after each epoch and stopped the training after 25 epochs.
Note that with more advanced techniques like AdaGrad~\cite{adagrad} and AdaDelta~\cite{adadelta} we did not achieve improvements over SGD with the described simple learning rate schedule. 
Since the learning schedule does not need a cross-validation set, we trained the final best model with the complete training data set.
Table \ref{tab:params} shows the hyper-parameters used for all the CNN models.

\begin{table}[ht]
\caption{{\it Hyper-parameters of sequential CNN}}
\label{tab:params}
  \vspace{2mm}
  \centerline{
\begin{tabular}{l|r}
 Parameters & Value \\
 \hline
 activation function & sigmoid\\
 number of features maps & 100 \\
 features map window & (50, 5) \\
 surrounding context & 3 \\
 context length (past or future) & 9 \\
 word embs & 50 \\
 \hline
 regularization & L2 \\
 L2 weight & 1e-7 \\
 %mini batch size & 1\\
 initial learning rate & 0.02 \\
\end{tabular}}
\end{table}

\subsection{Results}
We adopted the window approach proposed in \cite{nlp_2} as the baseline system.
Five left context words, five right context words and the current word form the input of a feed-forward neural network with one hidden layer with size 100.
We obtained an F1-score of 94.23\% and 94.14\% with this simple feed-forward network using ranking loss and hinge loss respectively.
Table \ref{tab:sCNN} summarizes the performance on the ATIS test set with different CNN architectural setups.
The results show that the context information from the past is more important than the future context. 
The future context, however, appears to provide meaningful information because their combination leads to better results.
Moreover, the comparison between two different kinds of combinations of previous and future context (concatenation vs. addition) suggests to not mix up the information using addition.
Finally, results in Table \ref{tab:sCNN} also reveal that using the ranking loss function proposed in \cite{ranking_loss} outperforms the hinge loss function.

\begin{table}[h]
\caption{{\it F1-score (\%) of uni vs. bi-directional sequential CNNs trained with two different ranking loss functions}}
\label{tab:sCNN}
\vspace{2mm}
\centerline{
\begin{tabular}{l|l|r}
Objectives & Methods & Score \\
\hline
Hinge loss & Words with surrounding context = 5 & 94.14 \\
Ranking loss & Words with surrounding context = 5 & 94.23 \\
\hline
Hinge loss & Past sequential CNN & 94.89 \\
 & Future sequential CNN & 93.04 \\
 & Bi-directional sequential CNN (add) & 94.78 \\ 
 & Bi-directional sequential CNN (concat) & 94.98  \\
\hline
Ranking loss & Past sequential CNN & 95.31 \\
& Future sequential CNN & 93.59 \\
& Bi-directional sequential CNN (add) & 95.19\\
& Bi-directional sequential CNN (concat) & \textbf{95.61} \\
\end{tabular}
}
\end{table}

\section{Analysis}
We performed analyses regarding the choice of context length, the impact of including the current word with its surrounding context and the most important detected n-grams.
\subsection{Context length}
First, the impact of the context length on the final performance was explored.
The number of parameters remained unchanged when reducing or increasing the context length.
Short context means information loss while a long context length potentially adds noise to the input of the model.
Table \ref{tab:contextLength} shows that F1-scores increased when increasing the context length from 5 up to 9. 
Increasing the context length to 10 and 11, however, decreased the results slightly but the F1-scores stayed quite stable around 95.5\%.
This confirms our hypothesis that a longer context adds noise to the input while the model is still able to extract the important information for slot prediction.
%\begin{figure}[th]
%\centering
%\includegraphics[width=.4\textwidth]{contextLength.png}
%\caption{Impact of context length}
%\label{fig:contextLength}
%\end{figure}
\begin{table}[th]
\caption{{\it Impact of the context length on the F1-score (\%)}}
\label{tab:contextLength}
\vspace{2mm}
\centerline{
\begin{tabular}{l|r|r|r|r|r}
Context length & 5 & 7 & 9 & 10 & 11 \\
 \hline
F1-score & 94.19 & 95.17 & 95.61 & 95.42 & 95.51 \\
\end{tabular}}
\end{table}
\vspace{-0.35cm}
\subsection{Surrounding context}
Table \ref{tab:surrounding} summarizes the F1-score without using the current word or with the current context with various lengths of the surrounding contexts.
The results revealed the strong impact of including the current word with its surrounding context into the CNN on the final F1-score.
Without paying attention to the current word, the F1-score dropped significantly to 92.01\%.
Successively adding the current word and increasing its surrounding contexts up to three left and three right neighbour words resulted in better performance.
Increasing the surrounding context to four, however, decreased F1-score. 
The best F1-score was obtained with three left and three right neighbour words.
\begin{table}[th]
\caption{{\it Impact of including the current word with surrounding context into the CNN on the F1-score (\%)}}
\label{tab:surrounding}
  \vspace{2mm}
  \centerline{
\begin{tabular}{l|r}
 Methods & F1-score \\
 \hline
Bi-directional sequential CNN (concat) &  \\
- current word & 92.01 \\
\hline
+ current word w/o context & 95.09 \\
+ surrounding context = 1 & 95.21 \\
+ surrounding context = 2 & 95.37 \\
+ surrounding context = 3 & 95.61 \\
+ surrounding context = 4 & 95.41 \\
\end{tabular}}
\end{table}
\vspace{-0.35cm}
\subsection{Most important n-grams}
We analyzed the most significant patterns for the four most frequent semantic slots in the test data.
For each of them, we present up to three n-grams which contributed the most to scoring the correctly classified test data points.
To compute the most important n-grams, we first detected the position of the maximum contribution to the dot product and traced it back to the corresponding feature map. 
Based on the max pooling, we were able to trace back and identify the n-grams which were used.
To create the results presented in Table \ref{tab:ngrams}, we ranked the n-grams which were selected as the most important features in all the sentences based on frequency and picked the most frequent ones.
Table \ref{tab:ngrams} shows that the model has learnt something meaningful for this task.
For example, a pattern such as \textit{flights from A to B} was used to predict \texttt{fromloc.city\_name} while the model only used \textit{A to B} or  \textit{to B} for \texttt{toloc.city\_name} prediction.
Other examples are patterns such as \textit{afternoon, evening} and \textit{night} which appeared quite frequently after \texttt{depart\_date.day\_name} and therefore are learnt as indicators.

\begin{table}[th]
\caption{{\it Most important n-grams for slot prediction}}
\label{tab:ngrams}
\vspace{2mm}
\centerline{
\begin{tabular}{l|r}
Slots & n-grams \\
\hline
\hline
fromloc.city\_name & \emph{'flights from washington dc to'} \\
                                   & \emph{'flights from ontario california to'} \\
                                   & \emph{'from toronto to san diego'} \\
                                   \hline
toloc.city\_name & \emph{'toronto to san diego'} \\
& \emph{'st. louis to burbank'} \\
\hline
depart\_date.day\_name & \emph{'afternoon sentence\_end'}\\
& \emph{'evening sentence\_end'}\\
& \emph{'night  sentence\_end'}\\
\hline
airline\_name & \emph{'northwest us air and united'} \\
& \emph{'show delta airlines flights from'} \\
\end{tabular}}
\end{table}

\vspace{-0.3cm}
\section{Comparison with state of the art}
\label{sec:comparison}
Table \ref{tab:compare} lists several previous results on the ATIS data set including our best results.
\begin{table}[th]
\caption{{\it Comparison with state-of-the-art results}}
\label{tab:compare}
  \vspace{2mm}
  \centerline{
\begin{tabular}{l|r}
 Methods & F1-score \\
 \hline
 CRF~\cite{rnn_slu_2} & 92.94\\
 simple RNN~\cite{rnn_slu_1} & 94.11 \\
 CNN~\cite{cnn_slu} & 94.35 \\
 LSTM~\cite{lstm_slu} & 94.85\\
 RNN-EM~\cite{rnn-mem} & 95.25 \\
 R-bi-RNN~\cite{rnn-rank} & 95.47 \\
 \hline
 R-bi-sCNN & \textbf{95.61} \\
\end{tabular}}
\end{table}
The proposed R-bi-sCNN outperforms the previously best ranking bi-directional RNN (R-bi-RNN).
A more detailed comparison with R-bi-RNN shows that R-bi-sCNN performed as well as R-bi-RNN on the frequent semantic slots but outperformed R-bi-RNN on the rare slots. 
For example, rare slots such as \texttt{toloc.country\_name}, \texttt{days\_code}, \texttt{period\_of\_day}, which appeared less than six times in the training data, were correctly predicted with the R-bi-sCNN model but not with R-bi-RNN .
%, while they could not be correctly predicted using R-bi-RNN.
\vspace{-0.1cm}
\section{Conclusions}
\label{sec:conclusions}
This paper explored convolutional neural networks for the slot filling task in spoken language understanding.
Our novel CNN architecture - bi-directional sequential CNN - takes into account the information from the past and the future with preserved order information and pays special attention to the current word with its surrounding contexts.
To train the model, we compared two different ranking objective functions.
Our findings revealed that not forcing the model to learn a pattern for \texttt{O} class is helpful to improve the final performance.
Finally, our bi-directional sequential CNN achieves state-of-the-art results with an F1-score of 95.61\% on the ATIS benchmark dataset without using any additional linguistic knowledge and resources.
As future work, we aim to evaluate the proposed model on other datasets (e.g. data presented in \cite{future_1, future_2}).
\vspace{-0.1cm}
\section{Acknowledgements}
This work was funded by the German Science Foundation (DFG), Sonderforschungsbereich 732 \textit{Incremental Specification in Context}, Project A8, at the University of Stuttgart.

\newpage
\eightpt
%\bibliographystyle{IEEEtran}
%\bibliography{mybib}

\end{document}